\documentclass[letterpaper]{article} % DO NOT CHANGE THIS
\usepackage[]{aaai25}  % DO NOT CHANGE THIS
\nocopyright
\usepackage{times}  % DO NOT CHANGE THIS
\usepackage{helvet}  % DO NOT CHANGE THIS
\usepackage{courier}  % DO NOT CHANGE THIS
\usepackage[hyphens]{url}  % DO NOT CHANGE THIS
\usepackage{graphicx} % DO NOT CHANGE THIS
\usepackage{natbib}  % DO NOT CHANGE THIS AND DO NOT ADD ANY OPTIONS TO IT
\usepackage{caption} % DO NOT CHANGE THIS AND DO NOT ADD ANY OPTIONS TO IT
\usepackage{algorithm}
\usepackage{algorithmic}
\usepackage{multirow}
\usepackage{booktabs}
\usepackage{amssymb}
\usepackage{subcaption}
\usepackage{xcolor}
\usepackage{newfloat}
\usepackage{listings}
\DeclareCaptionStyle{ruled}{labelfont=normalfont,labelsep=colon,strut=off} % DO NOT CHANGE THIS
\floatstyle{ruled}
\newfloat{listing}{tb}{lst}{}
\floatname{listing}{Listing}
\title{MultiMath: Bridging Visual and Mathematical Reasoning for Large Language Models}
\author{
    Shuai Peng\textsuperscript{\rm 1}, Di Fu, Liangcai Gao\textsuperscript{\rm 1}, Xiuqin Zhong\textsuperscript{\rm 2}, Hongguang Fu\textsuperscript{\rm 2}, Zhi Tang\textsuperscript{\rm 1}
}
\affiliations{
    \textsuperscript{\rm 1}Peking University\\
    \textsuperscript{\rm 2}University of Electronic Science and Technology of China\\
    pengshuaipku@pku.edu.cn, 
    fudi.01@bytedance.com,
    gaoliangcai@pku.edu.cn,
    zhongxiuqin@uestc.edu.cn,
    fuhongguang@uestc.edu.cn,
    tangzhi@pku.edu.cn
}

\usepackage{bibentry}
\begin{document}

\maketitle

\begin{abstract}
The rapid development of large language models (LLMs) has spurred extensive research into their domain-specific capabilities, particularly mathematical reasoning. However, most open-source LLMs focus solely on mathematical reasoning, neglecting the integration with visual injection, despite the fact that many mathematical tasks rely on visual inputs such as geometric diagrams, charts, and function plots. To fill this gap, we introduce \textbf{MultiMath-7B}, a multimodal large language model that bridges the gap between math and vision. \textbf{MultiMath-7B} is trained through a four-stage process, focusing on vision-language alignment, visual and math instruction-tuning, and process-supervised reinforcement learning. We also construct a novel, diverse and comprehensive multimodal mathematical dataset, \textbf{MultiMath-300K}, which spans K-12 levels with image captions and step-wise solutions. MultiMath-7B achieves state-of-the-art (SOTA) performance among open-source models on existing multimodal mathematical benchmarks and also excels on text-only mathematical benchmarks. Our model and dataset are available at {\textcolor{blue}{\url{https://github.com/pengshuai-rin/MultiMath}}}.
\end{abstract}


\section{Introduction}
\label{intro}

The rapid development of large language models (LLMs) has ushered in significant advancements in various domains, with a focus on specialized capabilities, particularly mathematical reasoning. Many domain-specific language models have primarily concentrated on mathematical reasoning in isolation \cite{metamath, wizardmath, math-shepherd, deepseekmath}, while neglecting the integration with visual reasoning. Simultaneously, general-purpose open-source multimodal large language models (MLLMs) \cite{liu2023llava, minigpt4} often lack specificity in vertical domains, resulting in a subpar performance in mathematical reasoning tasks.

Currently, domain-specific MLLMs for mathematical reasoning can be categorized into two types. The first, represented by G-LLaVA \cite{gllava} and AlphaGeometry \cite{alphageometry}, focuses on geometric problem solving (GPS) \cite{seo2015GEOS, sachan2017GEOS++, inter-gps, geodrl} but falls short in other multimodal mathematical reasoning tasks, such as function plot reasoning and scientific chart QA \cite{mathvista}. The second, represented by Math-LLaVA \cite{mathllava}, builds upon an existing open-source MLLM with math finetuning. However, it underperforms in text-only mathematical reasoning tasks \cite{gsm8k, math, cmath} due to the lack of large-scale pretraining on math corpora and the absence of chain-of-thought (CoT) reasoning capabilities. Consequently, there remains a notable gap in the availability of an open-source MLLM that excels across a broad spectrum of mathematical reasoning tasks.

\begin{figure}[t]
    \centering
    \includegraphics[width=1\linewidth]{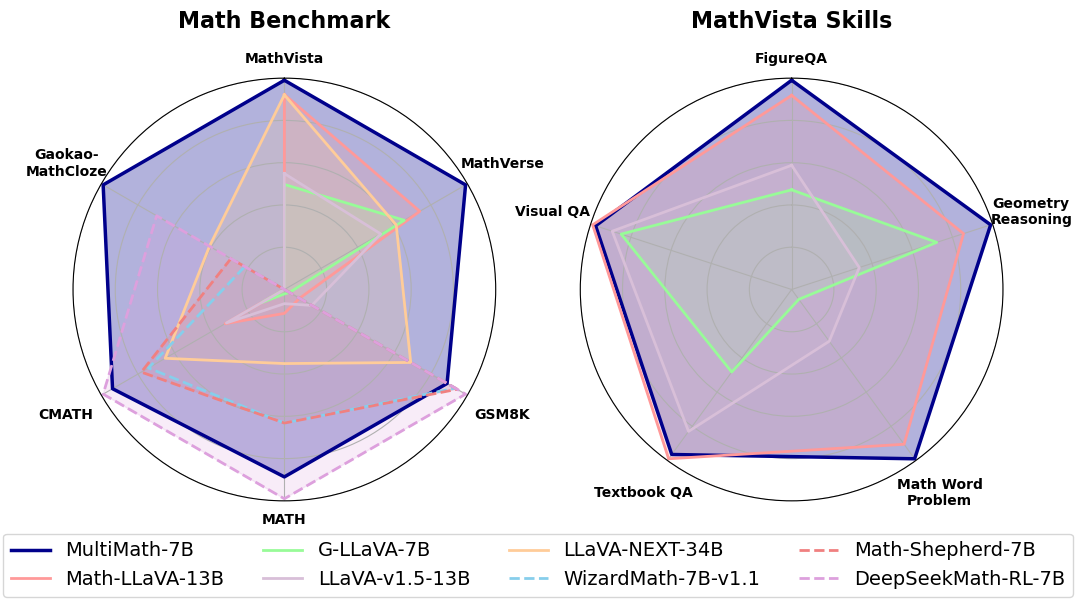}
    \caption{Comparison between MultiMath-7B and existing open-source MLLMs and Math LLMs across various math benchmarks and math skills. The data in the figure has been normalized.}
    \label{fig:spider}
\end{figure}

To bridge this gap, we introduce \textbf{MultiMath-7B}, a domain-specific multimodal large language model for mathematical reasoning. Unlike Math-LLaVA \cite{mathllava}, which directly applies math finetuning to existing MLLMs, we choose to build upon a well-trained math LLM as our foundation. We then enhance it with visual capabilities and align its visual and mathematical reasoning. This strategy leverages the reasoning abilities acquired from mathematical pretraining and extends them to the visual domain. MultiMath-7B employs DeepSeekMathRL-7B \cite{deepseekmath} as the foundation language model, augmented with a vision encoder and a multimodal adapter to enable visual capabilities. We adopt a multi-stage training process, progressively training the model’s visual alignment, visual dialogue, and visual reasoning abilities, ultimately bridging them with mathematical reasoning skills.

Another challenge in developing a math MLLM is the scarcity of multimodal alignment and instruction datasets in math domain. Existing open-source datasets typically focus on particular math subfields and lack visual-language alignment data and CoT-style instruction data. To address it, we construct \textbf{MultiMath-300K}, a \textit{multimodal}, \textit{multilingual}, \textit{multi-level} and \textit{multistep} mathematical reasoning dataset that encompasses a wide range of K-12 level mathematical problems. 
MultiMath-300K demonstrates three key strengths over existing multimodal math datasets Geo170K \cite{gllava} and MathV360K \cite{mathllava}: \textbf{Novelty}: the problems are not present in previously released datasets. \textbf{Diversity}: MultiMath-300K covers almost all K-12 grades, including a variety of math problem types such as arithmetic, algebra, geometry, function, algorithm, etc. \textbf{Comprehensiveness}: each problem is accompanied by an image caption for vision-language alignment training and a step-by-step solution for CoT instruction fine-tuning. The comparison of MultiMath-300K with Geo170K and MathV360K is shown in Table \ref{tab:dataset_comparison}.

Experimental results on mathematical reasoning tasks demonstrate that MultiMath-7B not only achieves SOTA performance among open-source models on multimodal mathematical benchmarks but also excels on text-only mathematical benchmarks. 
Notably, multimodal training has been shown to improve the model's performance on certain text-only mathematical reasoning tasks, suggesting that incorporating multimodal reasoning can enhance the language model's overall reasoning abilities.

The main contributions are summarized as follows:

\begin{itemize}
    \item We propose \textbf{MultiMath-7B}, a math MLLM that achieves SOTA performance among open-source models on multimodal mathematical benchmarks and excels in text-only mathematical reasoning tasks.
    \item  We constructed \textbf{MultiMath-300K}, a \textit{multimodal}, \textit{multilingual}, \textit{multi-level} and \textit{multistep} mathematical reasoning alignment and instruction dataset, covering a wide range of K-12 level mathematical problems.
    \item  We introduce a training framework for enhancing the multimodal capabilities of domain-specific models, preserving the original abilities while boosting multimodal performance. 
\end{itemize}

\section{Related Work}

\subsection{Multimodal Large Language Model}

Recent advancements in vision-language alignment and the maturation of large language model (LLM) have endowed LLMs with visual capabilities. Pioneering studies in vision-language alignment include CLIP \cite{CLIP} and BLIP \cite{BLIP}. CLIP aligns image and text semantic spaces through contrastive learning, while BLIP enhances visual-language understanding and generation by jointly training a vision encoder with a language model. Inspired by these models, researchers developed MLLMs such as Mini-GPT4 \cite{minigpt4} and LLaVA \cite{liu2023llava}, which leverage vision-language alignment training and instruction-tuning to enable LLMs to handle multimodal tasks. Recently, closed-source MLLMs like GPT-4V \cite{gpt4}, Gemini Pro \cite{gemini}, and Claude 3 \cite{claude3} have further pushed the boundaries of visual understanding capabilities. The typical training framework involves using pretrained vision encoders and language models, aligning them with visual caption data, and finally finetuning on instruction data for task-specific abilities.

Despite these advancements, there is still a significant gap in the development of domain-specific MLLMs, particularly in mathematical reasoning. This gap is due to the lack of an effective training framework for adapting math LLMs to multimodalities and the scarcity of multimodal alignment and instruction reasoning data. In this paper, we aim to address these issues.

\begin{table}[t]
    \centering
    \small
    \setlength{\tabcolsep}{2pt}
    \fontsize{8pt}{10pt}\selectfont
    \begin{tabular}{c|c|ccc|cc|c}
    \toprule
        \multirow{2}{*}{ \textbf{Dataset} } & \multirow{2}{*}{ \textbf{Original} } & \multicolumn{3}{c}{ \textbf{Task} } &  \multicolumn{2}{c}{\textbf{Data}} & \multirow{2}{*}{ \textbf{CoT} } \\
        \cline { 3 - 7 } & & GPS & MWP & FQA & Align & QA &  \\
        \midrule
        Geo170K &  & \checkmark & & & \checkmark & \checkmark & \checkmark \\
        MathV360K & & \checkmark & \checkmark & \checkmark & & \checkmark & \\
        MultiMath-300K & \checkmark & \checkmark & \checkmark & \checkmark & \checkmark & \checkmark & \checkmark \\
    \bottomrule
    \end{tabular}
    \caption{Comparison with existing multimodal math reasoning datasets Geo170K and MathV360K.}
    \label{tab:dataset_comparison}
    \vspace{-0.1in}
\end{table}

\subsection{Mathematical Reasoning}

Automated mathematical reasoning is a significant research area in artificial intelligence. It typically includes tasks such as mathematical word problems (MWP) \cite{wang2018mathdqn}, geometry problem solving (GPS) \cite{inter-gps}, and automatic theorem proving (ATP) \cite{chou1996automated}. The emergence of large language models (LLMs) has led to their dominance in numerous mathematical reasoning benchmarks, driven by their extensive pretraining and advanced comprehension and reasoning capabilities. Mathematical reasoning has increasingly garnered attention from researchers and has become an essential benchmark for assessing LLMs. Several specialized LLMs, such as MetaMath \cite{metamath}, Math-Shepherd \cite{math-shepherd}, WizardMath \cite{wizardmath}, and DeepSeekMath \cite{deepseekmath}, have been developed to address these tasks. Derived from general-purpose LLMs, these models are fine-tuned to strengthen their mathematical abilities. Open-source LLMs have shown strong performance on mathematical reasoning benchmarks, highlighting the potential of domain-specific models.

A challenge of mathematical reasoning lies in reasoning with visual injection, including geometry diagrams, scientific charts, function plots, etc. However, existing math MLLMs are either limited to geometric problems, as seen with G-LLaVA \cite{gllava}, or underperform in text-only mathematical reasoning tasks and lack chain-of-thought capabilities, such as Math-LLaVA \cite{mathllava}. To address these issues, we construct a novel, diverse, and comprehensive multimodal math reasoning dataset, including visual-language alignment data and step-by-step reasoning instructions. We use this dataset to train MultiMath-7B, filling the gap in open-source math MLLMs.

\section{Dataset}

\begin{figure}[t]
    \centering
    \includegraphics[width=1\linewidth]{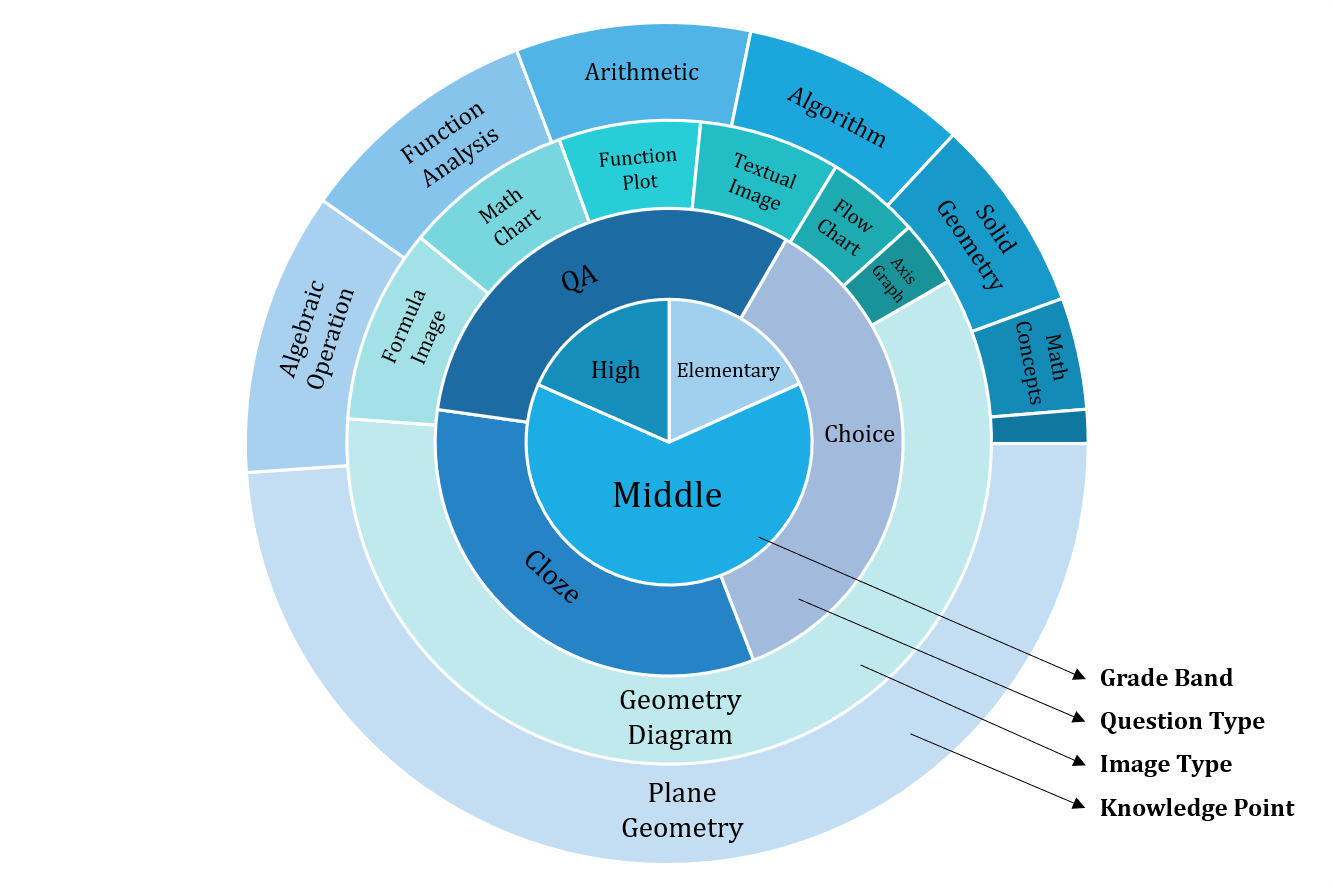}
    \caption{Statistics of MultiMath-300K, where each ring corresponds to an individual statistical dimension.}
    \label{fig:dataset_stats}
    \vspace{-0.1in}
\end{figure}

\begin{figure}[t]
    \centering
    \includegraphics[width=1\linewidth]{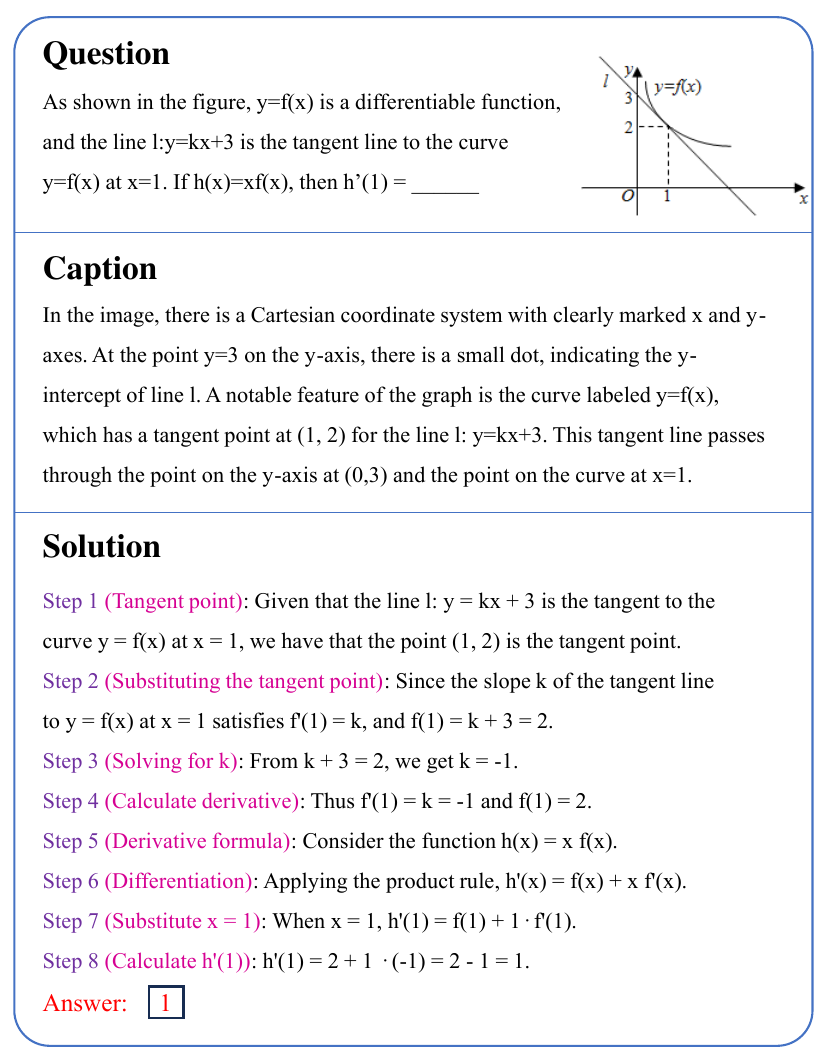}
    \caption{A data sample from MultiMath-300K, including the statement, image, caption, and solution, all in English. A complete sample also contains the Chinese statement, caption, solution, and other fields.}
    \label{fig:data_case}
    \vspace{-0.1in}
\end{figure}

In this section, we introduce the MultiMath-300K dataset, with a focus on the construction process.

\subsection{Overview}

MultiMath-300K comprises 298,670 mathematical problems, with 290,227 in the training set and 8,443 in the validation set. Each problem features an image and a statement in both English and Chinese. Covering all K-12 education levels, MultiMath-300K includes knowledge points such as arithmetic, algebra, mathematical concepts, plane geometry, solid geometry, function analysis, and algorithm derivation. Figure \ref{fig:dataset_stats} illustrates these statistics in a pie chart.

In addition to the problem data, MultiMath-300K includes vision-language alignment data and step-by-step solution instructions. The alignment data details the image for vision-language alignment training. The instruction data provides step-by-step reasoning solutions, with each step featuring an ID, name, and content, culminating in a final answer marked in \textit{boxed}. Figure \ref{fig:data_case} presents a data sample of the English part.

\begin{figure*}[t]
    \centering
    \includegraphics[width=1\linewidth]{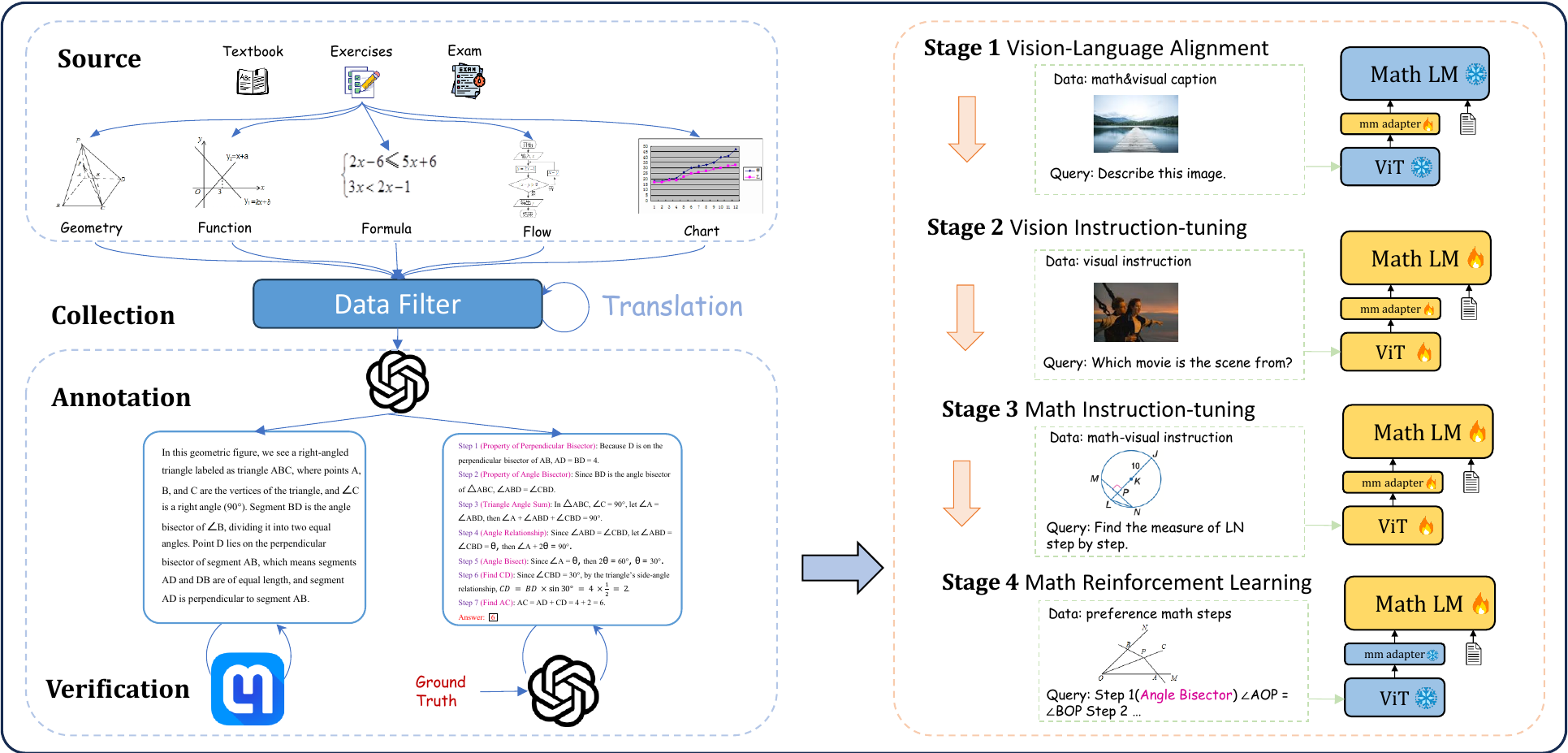}
    \caption{An illustration of the dataset construction and model training process. We collected problem sets from textbooks, exercises, and exams, and utilized GPT-4o for annotation and verification, producing the MultiMath-300K dataset for model training. The model's training is illustrated on the right, detailing the data types and training modules at each stage.}
    \label{fig:flow}
    \vspace{-0.1in}
\end{figure*}

\begin{figure}[t]
    \centering
    \includegraphics[width=1\linewidth]{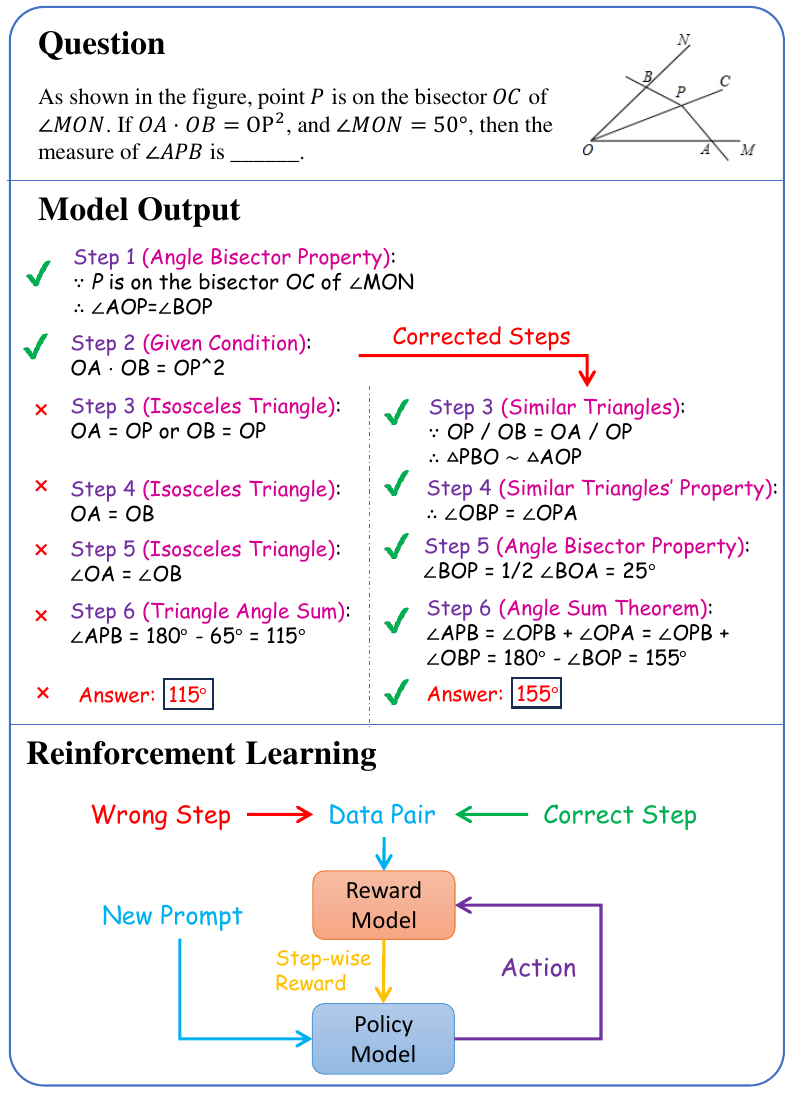}
    \caption{An illustration of RL data construction and training. Leveraging step-wise reasoning, GPT-4o identifies and corrects errors, generating preference data for training the reward model, which then guides reinforcement training with step-level rewards for error correction.}
    \label{fig:rl}
    \vspace{-0.1in}
\end{figure}
\subsection{Dataset Construction}

Here we outline the dataset construction process, encompassing collection, annotation, and verification, as illustrated on the left side of Figure \ref{fig:flow}.

\paragraph{Source Data} To ensure novelty, we collect mathematical problems from K-12 textbooks, exercises, and exams with authorization from the data providers. Selection criteria include (1) Completeness, requiring each problem to include the question title, details, solution, and standard answer. (2) Multimodality, with each problem featuring exactly one image. (3) Clarity, excluding images that are too small or blurry. This process initially yields 390,000 raw problems. We then use GPT-4-1106-preview to translate the problem texts, resulting in bilingual descriptions in both Chinese and English.

\paragraph{Alignment Data} The alignment data serve two purposes: (1) facilitating vision-language alignment training for multimodal models and (2) enabling language models to address multimodal problems through text-only descriptions. We utilized GPT-4o-2024-05-13 to generate bilingual captions for images in both Chinese and English. To address GPT-4o's limitations in OCR accuracy, we employed MathPix \footnote{https://mathpix.com/} to verify and correct OCR results for formula and textual images.

\paragraph{Instruction Data} Chain-of-thought (CoT) \cite{cot} reasoning has proven effective in enhancing LLM's mathematical reasoning abilities. To effectively utilize CoT reasoning, step-by-step instructional data is essential for model training, as it supports precise tracking of reasoning errors and enables fine-grained tuning. Therefore, our objective is to construct multistep reasoning data. We employed GPT-4o-2024-05-13 and GPT-4-1106-preview for annotation, conducting multiple rounds of refinement to ensure high-quality results, as follows:
\begin{itemize}
    \item[] \textbf{Round 1:} Generate step-by-step reasoning chains using GPT-4o, with detailed solutions from the original data as the hint.
    \item[] \textbf{Round 2:} Evaluate GPT-4o’s reasoning chains against the standard answers. If inconsistencies are found, require GPT-4o to revise the reasoning steps.
    \item[] \textbf{Round 3:} Submit GPT-4o’s responses and the standard answers to GPT-4 for verification, and retain only the correct answers.
\end{itemize}
Following these rounds of refinement, we compiled 300K problems to create MultiMath-300K. For further details on our dataset and the prompts used in its construction, please refer to the Appendix.

\section{Model Training}

In this section, we introduce the proposed mathematical multimodal large language model, MultiMath-7B. Compared to existing open-source mathematical MLLMs, our model offers three main advantages: (1) it tackles a broad spectrum of multimodal mathematical reasoning tasks, (2) it utilizes chain-of-thought (CoT) for detailed step-by-step reasoning, and (3) it maintains strong performance in text-only mathematical reasoning tasks. The appendix details the model training settings.

\subsection{Model Architecture}

MultiMath-7B is built upon the LLaVA architecture \cite{liu2023improvedllava} and integrates three primary components: a vision encoder, a multimodal adapter, and a language model. The vision encoder is initialized with \textit{openai/clip-vit-large-patch14-336}, which supports a 336×336 image resolution to effectively capture and recognize small text and mathematical symbols. The multimodal adapter is a two-layer MLP, initialized randomly. The language model is based on DeepSeekMath-RL \cite{deepseekmath}, a leading open-source 7B model in math reasoning.

\subsection{Training Stage}

Here we detail the training process of MultiMath-7B, presenting a novel framework for enhancing the multimodal capabilities of domain-specific LLMs. The overview is depicted on the right side of Figure \ref{fig:flow}. The training is structured into four stages, each addressing distinct aspects: vision-language alignment, visual instruction-tuning, math instruction-tuning, and finally, math process-supervised reinforcement learning. This sequential approach enables the model to extend its mathematical reasoning ability to the visual domain.

\paragraph{Vision-Language Alignment} In this stage, we focus on aligning the vision encoder and language model, enabling the latter to integrate visual information, which it has not previously processed. We train only the multimodal adapter while keeping the other modules freezed. Considering the potential lack of expertise of the initial vision encoder in mathematical content, we mix LLaVA-Pretrain \cite{liu2023improvedllava} dataset with domain-specific data from MultiMath300K alignment data and geo170k-align \cite{gllava}. The model is then trained for one epoch to align visual and language features within the mathematical domain.

\paragraph{Vision Instruction-tuning} This stage aims to enhance the model's visual comprehension and question-answering abilities. Although the model can now interpret visual information after stage 1, it still struggles with various visual tasks. To address this, we train all model components for two epochs using the LLaVA-Instruction \cite{liu2023improvedllava} dataset, which focuses on improving visual comprehension and question-answering capabilities.

\paragraph{Math Instruction-tuning} In this stage, we focus on extending mathematical reasoning capabilities to visual data, emphasizing chain-of-thought (CoT) reasoning. The CoT training is primarily driven by the MultiMath300K-instruction dataset. Additionally, we incorporate two open-source multimodal mathematical QA datasets, Geo170k-qa \cite{gllava} and MathV360k \cite{mathllava}, to further enhance the model’s performance. This combined training, conducted over two epochs, refines all model components and results in the instruction model.

\paragraph{Process-supervised Reinforcement Learning} This stage aims to correct errors at the step level during reasoning. Unlike supervised fine-tuning (SFT) in stage 2 and 3, reinforcement learning (RL) enhances the model’s ability to identify and correct reasoning errors more effectively. We use MultiMath300K-val, GSM8K-train \cite{gsm8k}, MATH-train \cite{math}, and CMATH-train \cite{cmath} for PPO \cite{ppo} training. The RL training process, illustrated in Figure \ref{fig:rl}, is summarized as follows:
\begin{enumerate}
    \item Given a mathematical problem, the instruction model performs chain-of-thought (CoT) reasoning and generates a result consisting of multiple reasoning steps.
    \item Given the standard answer and the model output from the previous step, GPT-4o accesses the correctness of the response. If incorrect, it identifies the step where the error occurred and regenerates the correct solution from that step.
    \item The correct and incorrect answers from the previous step form a paired preference dataset, used to train a reward model initialized from the instruction model.
    \item The reward model assigns a reward score to each reasoning step (action) generated by the policy model, supervising the policy model's gradient descent.
\end{enumerate}
This process results in the final RL-enhanced model. We will discuss the performance improvements from reinforcement learning in the Discussion section.

\begin{table*}[t]
    \centering
    \small
    \begin{tabular}{l|c|ccccc|c|ccccc}
        \toprule
        \multirow{2}{*}{ \textbf{Model} } & \multicolumn{6}{c}{ \textbf{MathVista} } &  \multicolumn{6}{c}{\textbf{MathVerse}} \\
        \cline { 2 - 13 } & \textbf{ALL} & FQA & GPS & MWP & TQA & VQA & \textbf{ALL} & TD & TL & VI & VD & VO \\
        \midrule
        \multicolumn{13}{c}{ \textit{Heuristics Baselines} } \\
        \midrule
        Random & 17.9 & 18.2 & 21.6 & 3.8 & 19.6 & 26.3 & 12.4 & 12.4 & 12.4 & 12.4 & 12.4 & 12.4  \\
        Human & 60.3 & 59.7 & 48.4 & 73.0 & 63.2 & 55.9 & 64.9 & 71.2 & 70.9 & 41.7 & 68.3 & 66.7 \\
        \midrule
        \multicolumn{13}{c}{ \textit{Closed-Source MLLMs} } \\
        \midrule
        GPT-4o \cite{gpt4} & \textbf{63.8} & - & - & - & - & - & - & - & - & - & - & - \\
        GPT-4V \cite{gpt4} & 49.9 & 43.1 & \textbf{50.5} & \textbf{57.5} & \textbf{65.2} & 38.0 & \textbf{54.4} & \textbf{63.1} & \textbf{56.6} & \textbf{51.4} & \textbf{50.8} & \textbf{50.3} \\
        Gemini Pro \cite{gemini} & 63.9 & - & - & - & - & - & 35.3 & 39.8 & 34.7  & 32.0 & 36.8 & 33.3 \\
        Claude 3.5 \cite{claude3} & 67.7 & - & - & - & -  & - & - & - & - & - & - & - \\
        Qwen-VL-Plus \cite{qwenvl} &  43.3 & \textbf{54.6} & 35.5 & 31.2 & 48.1 & \textbf{51.4} & 21.3 & 26.0 & 21.2 & 18.5 & 19.1 & 21.8  \\
        \midrule
        \multicolumn{13}{c}{ \textit{Open-Source MLLMs} } \\
        \midrule
        mPLUG-Owl2-7B \cite{mplug-owl2} & 22.2 & 22.7 & 23.6 & 10.2 & 27.2 & 27.9 & 8.3 & 8.9 & 9.1 & 10.2 & 8.1 & 5.3 \\
        MiniGPT4-7B \cite{minigpt4} & 23.1 & 18.6 & 26.0 & 13.4 & 30.4 & 30.2  & 12.2 & 12.3 & 12.9 & 12.5 & 14.8 & 8.7 \\
        LLaVA-1.5-13B \cite{liu2023improvedllava} & 27.7 & 23.8 & 22.7 & 18.9 & 43.0 & 30.2 & 14.3 & 20.3 & 11.1 & 14.9 & 13.2 & 12.0  \\
        SPHINX-V2-13B \cite{sphinx} & 36.7 & \textbf{54.6} & 16.4 & 23.1 & 41.8 & \textbf{43.0} & 16.1 & 20.4 & 14.1 & 14.0 & 15.6 & \textbf{16.2} \\
        LLaVA-NeXT-34B \cite{liu2024llavanext} & 46.5 & - & - & - & - & - & 16.6 & 24.8 & 12.0 & 18.2 & 13.9 & 14.1 \\
        % 45.0 & 33.2 & 30.1 & 55.1 & 35.8
        G-LLaVA-7B \cite{gllava} & 25.1 & 19.1 & 48.7 & 3.6 & 25.0 & 28.7 & 17.8 & 24.9 & 22.1 & 18.0 & 15.2 & 9.0 \\
        Math-LLaVA-13B \cite{mathllava} & 46.6 & 37.2 & 57.7 & 56.5 & \textbf{51.3} & 33.5 & 20.1 & 22.8 & 21.8 & 21.1 & 19.2 & 15.4 \\
        \midrule
        \textbf{MultiMath-7B} & \textbf{50.0} & 40.1 & \textbf{66.8} & \textbf{61.8} & 50.0 & 33.0 & \textbf{26.9} & \textbf{34.8} & \textbf{30.8} & \textbf{28.1} & \textbf{25.9} & 15.0 \\
        \bottomrule
    \end{tabular}
    \caption{Comparison with closed-source and open-source MLLMs on the testmini set of MathVista and MathVerse. }
    \label{tab:visual_benchmark}
    \vspace{-0.1in}
\end{table*}

\section{Experiment Results}

This section evaluates the performance of MultiMath-7B across various mathematical reasoning benchmarks, including visual and textual math reasoning tasks.

\subsection{Visual Math Benchmarks}

\paragraph{Datasets and Baselines} We select two representative multimodal mathematical reasoning datasets for evaluation: MathVista \cite{mathvista} and MathVerse \cite{mathverse}. MathVista assesses LLM' mathematical reasoning within visual contexts, while MathVerse presents more complex challenges in plane geometry, solid geometry, and functions. For evaluation, we utilize the provided prompts and perform zero-shot inference. Our baselines include closed-source MLLMs, open-source MLLMs, and two open-source MLLMs G-LLaVA \cite{gllava} and Math-LLaVA \cite{mathllava}.

\paragraph{Main Results} Table \ref{tab:visual_benchmark} presents the evaluation results of MathVista and MathVerse on the testmini dataset, including both closed-source and open-source MLLMs. MultiMath-7B sets a new state-of-the-art (SOTA) among open-source models for both benchmarks. Remarkably, despite having only 7B parameters, MultiMath-7B surpasses models with up to 34 billion parameters, demonstrating its exceptional performance in visual-mathematical reasoning tasks. Additionally, MultiMath-7B outperforms the closed-source Qwen-VL-Plus \cite{qwenvl} on both datasets, with its MathVista performance comparable to GPT-4V.

\paragraph{Subset Results} We also report the results on the subsets of MathVista and MathVerse: MathVista is divided into Figure QA, Geometry Problem Solving, Math Word Problem, Textbook QA, and Visual QA. MathVerse is categorized into Text Dominant, Text Lite, Vision Intensive, Vision Dominant, and Vision Only. MultiMath-7B notably excels across most subsets, significantly outperforming other MLLMs in Geometry Problem Solving and Math Word Problem tasks. It also leads open-source MLLMs in most MathVerse subsets, with the exception of the Vision Only category.

\subsection{Textual Math Benchmarks}

\begin{table}[t]
    \centering
    \scriptsize
    \setlength{\tabcolsep}{2pt}
    \fontsize{8pt}{10pt}\selectfont
    \begin{tabular}{l|cccccc}
        \toprule
        \multirow{3}{*}{ \fontsize{10pt}{8pt}\selectfont \textbf{Model} } & \multicolumn{2}{c}{\textbf{English}} & \multicolumn{2}{c}{\textbf{Chinese}} \\
        \cmidrule(lr){2-3} \cmidrule(lr){4-5}
        & \multirow{2}{*}{\fontsize{6pt}{8pt}\selectfont \textbf{GSM8K}} & \multirow{2}{*}{\fontsize{6pt}{8pt}\selectfont \textbf{MATH}} & \multirow{2}{*}{\fontsize{6pt}{8pt}\selectfont \textbf{CMATH}} & \fontsize{6pt}{8pt}\selectfont \textbf{Gaokao-} \\
        & & & & \fontsize{6pt}{8pt}\selectfont \textbf{MathCloze} \\
        \midrule
        \multicolumn{5}{c}{\textit{Closed-Source LLMs}} \\
        \midrule
        Gemini Ultra \cite{gemini} & \textbf{94.4} & \textbf{53.2} & - & - \\
        GPT-4 \cite{gpt4} & 92.0 & 52.9 & 86.0 & 22.0 \\
        GPT-3.5 \cite{gpt3} & 80.8 & 34.1 & 73.8 & 7.6 \\
        Gemini Pro \cite{gemini} & 86.5 & 32.6 & - & - \\
        \midrule
        \multicolumn{5}{c}{\textit{Open-Source Foundation LLMs}} \\
        \midrule
        Vicuna-7B \cite{vicuna} & 10.1 & 3.5 & 22.3 & 2.5 \\
        Mistral-7B \cite{mistral} & 40.3 & 14.3 &  \textbf{44.9} & 5.1 \\
        Llemma-7B \cite{llemma}  &  37.4 & 18.1 & 43.4 & \textbf{11.9} \\
        Llama-2-13B \cite{llama2} & 43.0 & - & - & - \\
        Llama-3-8B$^\dag$ \cite{llama3} & 79.6 & 30.0 & - & - \\
        Llama-3-70B$^\dag$ \cite{llama3} & \textbf{90.0} & \textbf{50.4} & - & - \\
        \midrule
        \multicolumn{5}{c}{\textit{Open-Source Math LLMs}} \\
        \midrule
        WizardMath-7B-v1.1 \cite{wizardmath} & 83.2 & 33.0 & 66.6 & 6.3 \\
        Math-Shepherd-7B \cite{math-shepherd} & 84.1 & 33.0 & 70.1 & 8.5 \\
        MetaMath-70B \cite{metamath} & 82.3 & 26.6 & 70.9 & - \\
        DeepSeekMath-7B \cite{deepseekmath} & \textbf{88.2} & \textbf{51.7} & \textbf{88.8} & \textbf{20.3} \\
        \midrule
        \multicolumn{5}{c}{\textit{Open-Source MLLMs}} \\
        \midrule
        G-LLaVA-7B \cite{gllava} & 2.5 & 1.1 & 11.1 & 0.8 \\
        Math-LLaVA-13B \cite{mathllava} & 7.4 & 5.9 & 29.0 & 0.0 \\ 
        LLaVA-1.5-7B \cite{liu2023improvedllava} & 13.4 & 3.5 & 28.4 & 0.0 \\
        LLaVA-NeXT-34B \cite{liu2024llavanext} & 61.5 & 18.3 & 58.4 & 11.9 \\
        \midrule
        \textbf{MultiMath-7B} & \textbf{79.2} & \textbf{46.3} & \textbf{84.2} & \textbf{28.8} \\ 
        \bottomrule
    \end{tabular}
    \caption{Results on textual math benchmarks. \dag: 8-shot for GSM8K and 4-shot for MATH.}
    \label{tab:text_benchmark}
    \vspace{-0.1in}
\end{table}

\paragraph{Datasets\&Baselines} We selected four representative textual mathematical reasoning datasets for evaluation: GSM8K \cite{gsm8k} and MATH \cite{math} in English, CMATH \cite{cmath} and Gaokao-MathCloze \cite{agieval} in Chinese. GSM8K and CMATH focus on elementary math, while MATH and Gaokao-MathCloze cover high school to university-level problems. We used MultiMath's chain-of-thought prompts and zero-shot inference to assess accuracy. For baseline comparisons, we include common closed-source LLMs, open-source foundation LLMs and math LLMs, as well as open-source MLLMs.

\paragraph{Results} Table \ref{tab:text_benchmark} presents the evaluation results on these benchmarks. While closed-source LLMs continue to lead in performance, open-source math LLMs closely follow. MultiMath-7B significantly outperforms 7B and 13B open-source foundation LLMs and MLLMs, but it slightly trails behind the top open-source math LLMs. Notably, despite a decline in text-only reasoning compared to DeepSeekMathRL-7B \cite{deepseekmath}, MultiMath-7B excels on the Gaokao-MathCloze dataset. This is attributed to its extensive training on Gaokao-style problems in MultiMath-300K, enhancing the model's capability to solve high school math questions. Additionally, G-LLaVA \cite{gllava} and Math-LLaVA \cite{mathllava} underperformed on text-only mathematical tasks, even compared to LLaVA-1.5-7B \cite{liu2023improvedllava} before its multimodal fine-tuning, indicating that their training is highly specialized for visual mathematical data and less effective for single-modal tasks.

\section{Discussion}

In this section, we explore the factors driving the model's performance, specifically, what contributes to the model's outcomes.

\paragraph{Visual Enhancement or Reasoning Boost?} The improvement of mathematical MLLM in multimodal math reasoning tasks compared to its foundation language models can be attributed to two main factors: (1) visual injection, which provides essential context for problem-solving;(2) finetuning on new math reasoning tasks, which boosts the model's reasoning ability in some aspects. To investigate this, we evaluate DeepSeekMathRL-7B on the text-only testmini set of MathVista (Table \ref{tab:visual_improvement}). Converting visual data into text allows the language model to solve multimodal math problems. With the same text-only inputs, MultiMath-7B achieved 9.1 points higher accuracy than DeepSeekMath-RL-7B, reflecting gains from reasoning boost alone. Inferencing with images further improves the performance by 4.3, indicating gains from visual injection. These findings suggest that while both factors contribute, reasoning boost plays a more substantial role. This supports our assertion that multimodal reasoning training can enhance reasoning abilities within a single modality.

\begin{table}[t]
    \centering
    \small
    \begin{tabular}{c|cc}
        \toprule
        \textbf{Model} & \textbf{Settings} & \textbf{MathVista} \\
        \midrule
        DeepSeekMath-RL-7B & text-only & 36.6 \\ 
        \midrule
        MultiMath-7B & text-only & 45.7 \\
        MultiMath-7B & with image &  50.0 \\
         \bottomrule
    \end{tabular}
    \caption{Comparison with DeepSeekMath-7B without multimodal-finetuned and MultiMath-7B on \textbf{text-only} testmini set of MathVista.}
    \label{tab:visual_improvement}
    \vspace{-0.1in}
\end{table}

\begin{figure}[t]
    \centering
    \includegraphics[width=0.8\linewidth]{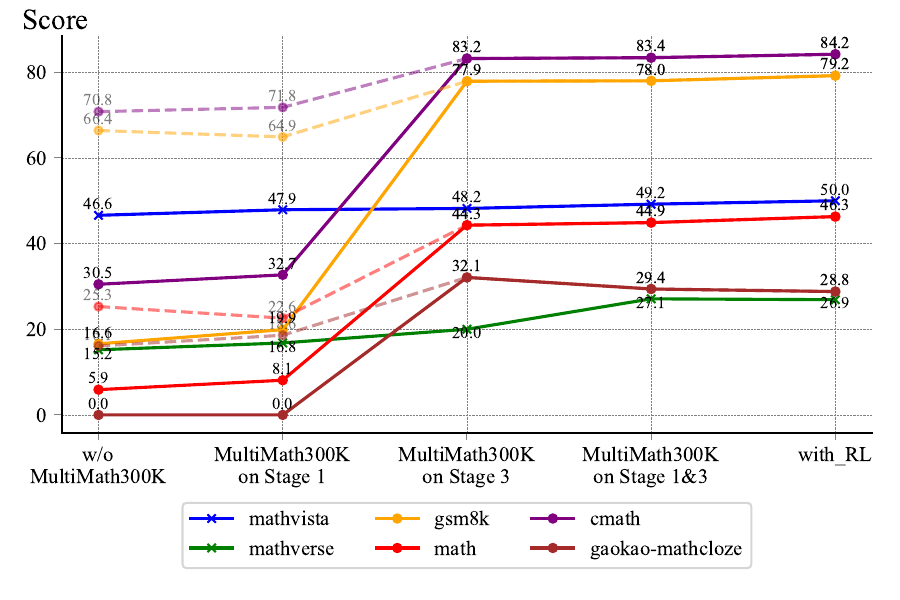}
    \vspace{-0.1in}
    \caption{Ablation studies of different training stages w/ or w/o MultiMath-300K and RL. The dashed lines denote without stage 3 instruction-tuning.}
    \label{fig:performance_on_settings}
    \vspace{-0.1in}
\end{figure}   
\begin{figure}[t]
    \centering
    \includegraphics[width=0.8\linewidth]{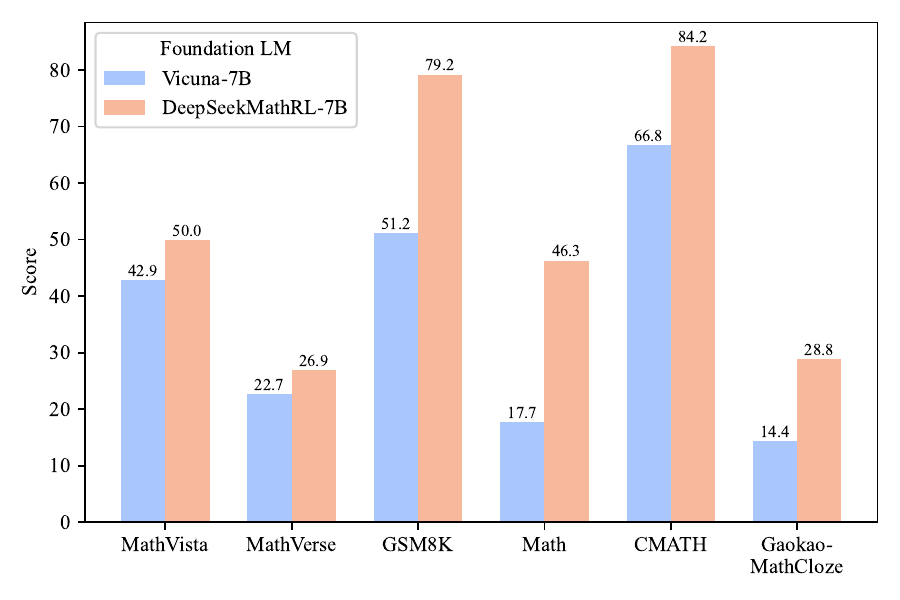}
    \vspace{-0.1in}
    \caption{Performance of different foundation models after MultiMath training.}
    \label{fig:ablation_foundationlm}
    \vspace{-0.1in}
\end{figure}

\paragraph{Contribution of Dataset} To assess the impact of the proposed MultiMath-300K dataset, we conducted ablation studies by excluding it from pretraining (Stage 1) and math instruction-tuning (Stage 3) and evaluate the models on six mathematical benchmarks (Figure \ref{fig:performance_on_settings}).
While MathV360K primarily boosted performance on MathVista, it significantly undermined the model's ability on textual math tasks. Incorporating MultiMath-300K during Stage 3 led to substantial improvements across nearly all benchmarks, highlighting its critical role in enhancing comprehensive mathematical reasoning. Additionally, Stage 1's math alignment training provided a modest performance gain.

\paragraph{Contribution of RL} Figure \ref{fig:performance_on_settings} also depicts the ablation results of stage 4 math reinforcement Learning. RL improved the model's performance on GSM8K, MATH, CMATH, and MathVista, but led to a decline on MathVerse and Gaokao-MathCloze. This aligns with expectations, as the RL training primarily used in-domain data from GSM8K and MATH, leading to better results on those benchmarks while negatively affecting out-of-domain datasets. This study confirms the viability of step-wise RL for multimodal math training, and future work could explore RL on larger, more diverse datasets to mitigate out-of-domain performance drops.

\paragraph{Contribution of Foundation LM} To assess how much of MultiMath-7B's performance attributed to its foundation model, DeepSeekMathRL, we retrained it using Vicuna-7B as a baseline. The results are shown in Figure \ref{fig:ablation_foundationlm}. DeepSeekMath outperforms Vicuna more significantly on textual benchmarks than visual benchmarks. This suggests the gains on visual tasks stem mainly from multimodal training rather than the language model itself. Additionally, compared to Table \ref{tab:text_benchmark}, Vicuna’s improvements after MultiMath training support the hypothesis that multimodal reasoning training enhances overall mathematical reasoning abilities.

\section{Conclusion}

In this paper, we introduce \textbf{MultiMath-7B}, a multimodal math large language model that bridges the gap between visual and mathematical reasoning. We also construct a multimodal math dataset \textbf{MultiMath-300K}, which spans K-12 levels and includes image captions and step-wise solutions. MultiMath-7B achieves SOTA performance among open-source models on existing multimodal mathematical benchmarks and also excels on text-only mathematical reasoning datasets. Future work will focus on expanding the model's training with diverse datasets across multiple domains and modalities to overcome its current limitations.

\bibliography{aaai25}

\newpage
\clearpage
% \documentclass[letterpaper]{article} % DO NOT CHANGE THIS
% \usepackage[submission]{aaai25}  % DO NOT CHANGE THIS
% \usepackage{times}  % DO NOT CHANGE THIS
% \usepackage{helvet}  % DO NOT CHANGE THIS
% \usepackage{courier}  % DO NOT CHANGE THIS
% \usepackage[hyphens]{url}  % DO NOT CHANGE THIS
% \usepackage{graphicx} % DO NOT CHANGE THIS
% \urlstyle{rm} % DO NOT CHANGE THIS
% \def\UrlFont{\rm}  % DO NOT CHANGE THIS
% \usepackage{natbib}  % DO NOT CHANGE THIS AND DO NOT ADD ANY OPTIONS TO IT
% \usepackage{caption} % DO NOT CHANGE THIS AND DO NOT ADD ANY OPTIONS TO IT
% \frenchspacing  % DO NOT CHANGE THIS
% \setlength{\pdfpagewidth}{8.5in} % DO NOT CHANGE THIS
% \setlength{\pdfpageheight}{11in} % DO NOT CHANGE THIS
% %
% % These are recommended to typeset algorithms but not required. See the subsubsection on algorithms. Remove them if you don't have algorithms in your paper.
% \usepackage{algorithm}
% \usepackage{algorithmic}
% \usepackage{multirow}
% \usepackage{booktabs}
% \usepackage{float}
% \renewcommand{\floatpagefraction}{.8}
% \renewcommand{\textfraction}{.1}

% \begin{document}
\section{Appendix}

\subsection{Dataset}

\paragraph{Source and Privacy} The math problems in MultiMath-300K are sourced from Xuekubao\footnote{http://test.xuekubao.com/}'s K12 question bank, which is collected from math textbooks, exercises, and exam questions. We purchased usage rights for the question bank and obtained permission for research purposes. During the filtering process, we removed any questions involving students' privacy. We also used an n-gram strategy to compare the data with existing mathematical reasoning datasets and filtered out duplicate questions to ensure the novelty of the dataset.

\begin{figure}[h]
    \centering
    \includegraphics[width=1\linewidth]{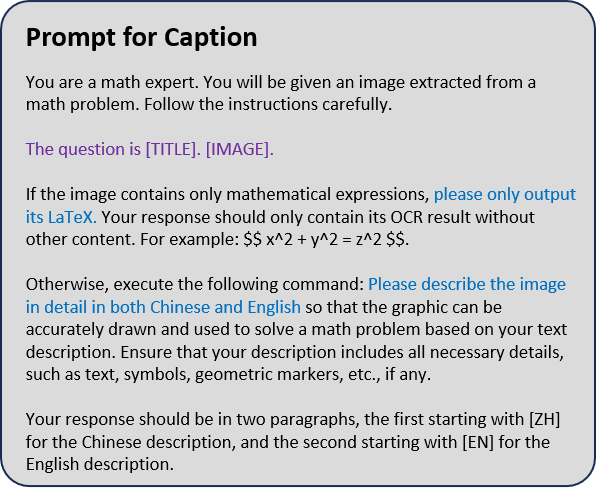}
    \caption{Prompt for the caption.}
    \label{fig:prompt_for_caption}
\end{figure}

\begin{figure}[h]
    \centering
    \includegraphics[width=1\linewidth]{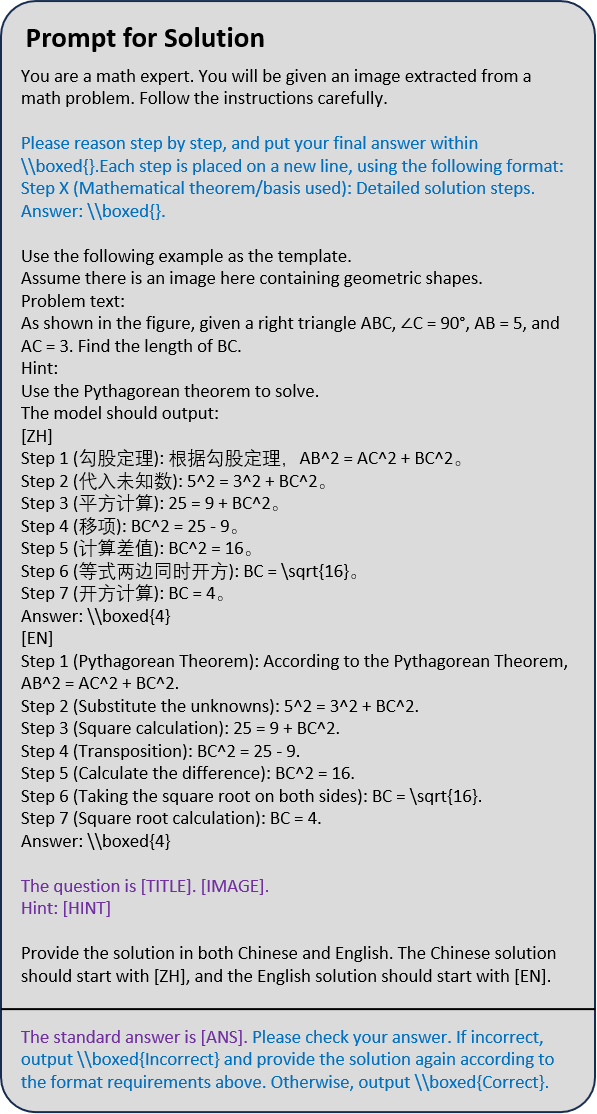}
    \caption{Prompt for the solution.}
    \label{fig:prompt_for_solution}
\end{figure}

\begin{figure}[h]
    \centering
    \includegraphics[width=1\linewidth]{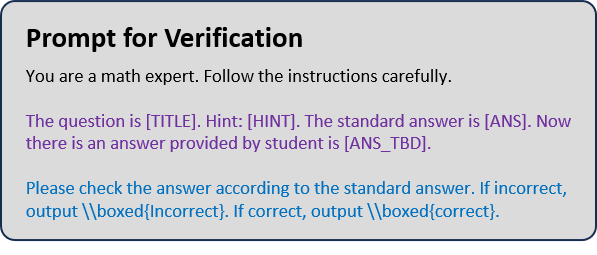}
    \caption{Prompt for verification.}
    \label{fig:prompt_for_verification}
\end{figure}

\paragraph{Prompt} We use GPT-4o-2024-05-13 to annotate the image captions and problem solutions. We present the prompts used in the dataset construction, including prompts for the caption (Figure \ref{fig:prompt_for_caption}), solution (Figure \ref{fig:prompt_for_solution}), and verification (Figure \ref{fig:prompt_for_verification}). In these figures, the texts in blue are instructions, and in purple are the input question information. We use these prompts to generate Chinese and English captions and solutions using GPT-4o-2024-05-13 (caption and solution) and GPT-4-1106-preview (verification).

\paragraph{Format}  We include one thousand data examples of MultiMath-300K in data appendix to demonstrate the data format. The complete dataset has been released on Hugging Face. 

\newpage
\clearpage

\begin{figure}[t]
    \centering
    \includegraphics[width=2\linewidth]{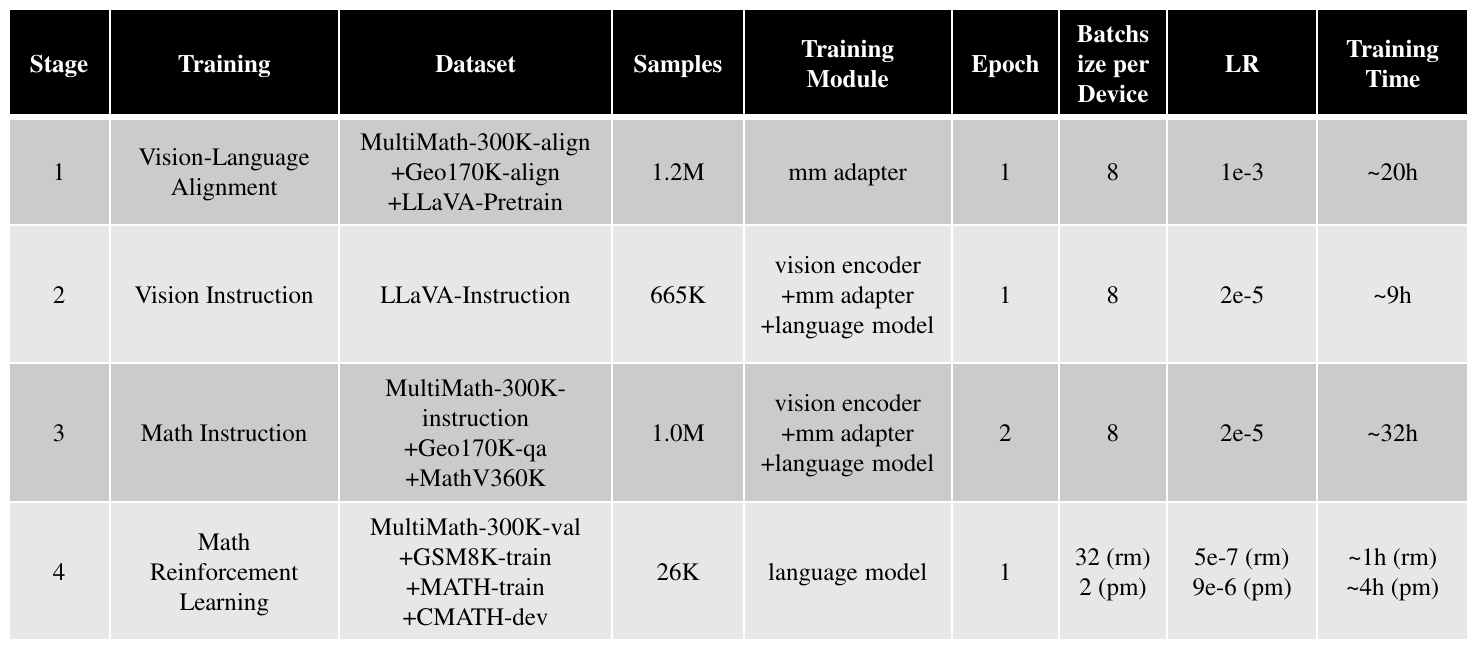}
    \caption{}
    \label{fig:training_stage_table}
\end{figure}

\subsection{Model Training}

Here we detail the datasets and settings used on each training stage in Figure \ref{fig:training_stage_table}. All the experiments were conducted on 8 NVIDIA A100-80GB GPUs with the random seed 42. For more implementation details, please refer to our code appendix. The model weights have been released on Hugging Face.

\subsection{Model Inference}

We inferred our model as well as other MLLMs with the settings of \textit{temperature: 0.2, top\_p: None, num\_beams: 1, max\_new\_tokens: 1024}. We evaluated three times on a task for each model and obtained the average score as the final accuracy.

% \end{document}

\end{document}